# Adaptive Cubic Regularized Second-Order Latent Factor Analysis Model

*Jialiang Wang[1], Junzhou Wang[1], Xin Liao [1*]*

[1]*College of Computer and Information Science, Southwest University, Chongqing, China*
**lxchat26@gmail.com*

**Keywords**: HIGH-DIMENSIONAL AND INCOMPLETE, LATENT FACTOR ANALYSIS, CUBIC REGULARIZED NEWTON, SECOND-ORDER OPTIMIZATION, HESSIAN-FREE OPTIMIZATION

**Abstract**

High-dimensional and incomplete (HDI) data, characterized by massive node interactions, have become ubiquitous across various real-world applications. Second-order latent factor models have shown promising performance in modeling this type of data. Nevertheless, due to the bilinear and non-convex nature of the SLF model's objective function, incorporating a damping term into the Hessian approximation and carefully tuning associated parameters become essential. To overcome these challenges, we propose a new approach in this study, named the adaptive cubic regularized second-order latent factor analysis (ACRSLF) model. The proposed ACRSLF adopts the two-fold ideas: 1) self-tuning cubic regularization that dynamically mitigates non-convex optimization instabilities; 2) multi-Hessian-vector product evaluation during conjugate gradient iterations for precise second-order information assimilation. Comprehensive experiments on two industrial HDI datasets demonstrate that the ACRSLF converges faster and achieves higher representation accuracy than the advancing optimizer-based LFA models.

## 1   Introduction

High-dimensional and incomplete (HDI) data are prevalent in big data scenarios characterized by extensive node interactions, such as a recommender system (RecSys) [1-20]. In RecSys, these interactions can be represented as a rating matrix, where each row represents a user, each column represents an item, and each element $r_{u,i}$ denotes the rating given by user $u$ to item $i$ [21-39]. With the rapid growth in the scale of users and items, individual users cannot rate all items, nor can all items be rated by every user. Consequently, the rating matrix becomes HDI. However, this HDI structure retains substantial valuable information.

The latent factor analysis (LFA) model has been demonstrated as an effective approach for extracting latent features from HDI matrices [41-60]. The LFA method assumes that user and item characteristics can be mapped into a shared low-rank latent factor space, enabling the approximation of HDI rating matrices through matrix factorization [61-80].

Mainstream LFA models predominantly employ first-order optimization algorithms, *e.g.*, momentum-based stochastic gradient descent, for objective function minimization [81-104]. To enhance low-rank representation capability, advanced implementations utilize second-order methods such as Hessian-free optimization [35, 38-40, 106]. Given the bilinear and non-convex nature of LFA's optimization landscape, the Hessian matrix in the second-order latent factor analysis (SLF) model may lose positive definiteness. This necessitates the widespread application of damped Gauss-Newton approximations for Hessian matrix estimation. Nevertheless, existing damping parameter configurations require manual mutual adjustment rather than automated epoch-wise self-adaptation.

To address this critical issue, we present a novel adaptive cubic regularized second-order latent factor analysis, ACRSLF for short. The proposed ACRSLF model with the following two-fold ideas:

1) The proposed ACRSLF model incorporates adaptive cubic regularization as a damping, ensuring positive definiteness of the curvature matrix while enabling self-adaption.

2) We develop a computationally efficient framework that embeds second-order information through iterative multi-Hessian-vector product calculations during conjugate gradient (CG), achieving gradient direction scaling through nonlinear preconditioning.

The experimental results on two RecSys datasets indicate that the novel ACRSLF model surpasses existing state-of-the-art LFA approaches.

This paper is organized as follows: Section II introduces theoretical preliminaries. Section III details the ACRSLF framework. Section IV presents experimental results. Section V draws the conclusion.

## 2. Preliminaries

*2.1 Problem Statement*

*Definition 1: (HDI Matrix)* Let $U$ denote a set of users and $I$ a set of items. The user-item interaction data can be organized into rating matrix $\mathbf{R} \in \mathbb{R}^{|U| \times |I|}$, where each element $r_{u,i}$ represents the rating provided by user $u \in U$ to item $i \in I$. Denote the set of observed entries $R_K$ and the unobserved ones as $R_M$. The matrix $\mathbf{R}$ is defined as a HDI matrix if the number of observed entries $|R_K|$ is significantly smaller than the number of missing entries $|R_M|$.

*Definition 2. (LFA Model)* Given sets of users $U$, items $I$, and a collection of known interactions $R_K$, the LFA model approximates the full interaction matrix $\mathbf{R}$ through low-rank matrix factorization. Specifically, it seeks to reconstruct the $\mathbf{R}$ by $\hat{\mathbf{R}} \approx \mathbf{Y}_U \mathbf{Y}_I^T$, where $\mathbf{Y}_U \in \mathbb{R}^{|U| \times f}$ and $\mathbf{Y}_I \in \mathbb{R}^{|I| \times f}$ are trainable embedding matrices with latent dimension $f$. The estimated rating $\hat{r}_{u,i} \approx y_u y_i^T$, where $y_u$ and $y_i$ denote the $u$-th and $i$-th row vectors of $\mathbf{Y}_U$ and $\mathbf{Y}_I$, respectively.

To learn optimal latent factor representations for users and items, we formulate the optimization problem as minimizing the discrepancy between the observed rating matrix $\mathbf{R}$ and its approximation $\hat{\mathbf{R}}$. As established in prior LFA studies [70-80], the objective function is constructed as follows:

$$E(\mathbf{y}) = \frac{1}{2} \sum_{r_{u,i} \in R_K} \left( \left( r_{u,i} - y_u y_i^T \right)^2 + \lambda \left( \|y_u\|^2 + \|y_i\|^2 \right) \right)$$
$$= \frac{1}{2} \left( \sum_{r_{u,i} \in R_K} \left( r_{u,i} - \sum_{d=1}^{f} y_{u,d} y_{i,d} \right)^2 + \lambda \sum_{d=1}^{f} \left( y_{u,d}^2 + y_{i,d}^2 \right) \right), \quad (1)$$
$$s.t. \ \forall u \in U, i \in I, d \in \{1,...,f\},$$

where $\mathbf{y} = vec(\mathbf{Y}_U, \mathbf{Y}_I) \in \mathbb{R}^{(|U| \times |I|) \times f}$, with $vec(\cdot)$ representing the vectorization operation. $E(\mathbf{y})$ denotes the objective function w.r.t parameter $\mathbf{y}$, $\lambda$ controls the strength of the regularization term. The notation $\|\cdot\|$ refers to the $L_2$-norm. The symbols $y_{u,d}$ and $y_{i,d}$ indicate the $d$-th element of the latent factors $y_u \in \mathbb{R}^f$ and $y_i \in \mathbb{R}^f$, respectively.

*2.2 Second-Order Latent Factor Analysis Model*

Generally, second-order techniques aim to minimize the function as follows [8, 17]:

$$\arg\min_{\Delta \mathbf{y}} \mathbf{g}_E(\mathbf{y}) + \left( \mathbf{H}_E(\mathbf{y}) + \gamma \mathbf{I} \right) \Delta \mathbf{y} \leq \tau \mathbf{1}, \quad (2)$$

where $\mathbf{g}_E(\mathbf{y}) \in \mathbb{R}^{(|U| \times |I|) \times f}$ is the gradient of $E(\mathbf{y})$, $\Delta \mathbf{y} \in \mathbb{R}^{(|U| \times |I|) \times f}$ is the update vector. Tolerance $\tau$ defines a convergence threshold, $\mathbf{1} \in \mathbb{R}^{(|U| \times |I|) \times f}$ is a vector filled with ones. The curvature matrix $\mathbf{H}_E(\mathbf{y})$, often chosen as the Hessian or its approximation, is regularized using the identity matrix $\mathbf{I}$ scaled by a damping factor $\gamma$. Both $\mathbf{H}_E(\mathbf{y})$ and $\mathbf{I}$ are square matrices with dimensions matching the size of the latent factor vector $\mathbf{y}$. The regularization term $\gamma \mathbf{I}$ helps stabilize the optimization by combining curvature information with gradient-based updates, and ensures that the modified matrix remains positive definite.

*2.3 Cubic Regularized Newton*

As established in [11, 12], the cubic regularized Newton method provides an effective framework for non-convex optimization, particularly addressing the challenge of escaping strict saddle points. This second-order optimization scheme iteratively solves the following subproblem at each step $t$:

$$T_M(\Delta \mathbf{y}) = \arg\min_{\Delta \mathbf{y}} E(\mathbf{y}) + \Delta \mathbf{y}^T \mathbf{g}_E(\mathbf{y}) + \frac{1}{2} \Delta \mathbf{y}^T \mathbf{H}_E(\mathbf{y}) \Delta \mathbf{y} + \frac{M}{6} \|\Delta \mathbf{y}\|^3, \quad (3)$$

where $M$, a cubic regularized parameter, ensures controlled step sizes while maintaining convergence guarantees. The control parameter $M > 0$ adaptively modulates the local model's curvature approximation accuracy.

## 3 Method

*3.1 Mapping Trick*

The mathematical analysis process of the second-order-based LFA model can be streamlined by two-fold modifications [17]: 1) projecting the bilinear interaction $y_u y_i^T$ through a hidden function $h(\cdot)$, e.g., $y_u y_i^T = h_{(u,i)}(\mathbf{y})$, and 2) removing the $L_2$ regularization constraints.

This refined formulation yields the following loss function $L(h(y))$ for the optimized LFA model with hidden mapping trick:

$$L(h(y)) = \frac{1}{2} \sum_{r_{u,i} \in R_K} \left( r_{u,i} - h_{(u,i)}(y) \right)^2. \tag{4}$$

*3.2 Gauss-Newton Approximation*

The Gauss-Newton matrix has proven to be efficient in approximating the Hessian matrix of a non-convex function [17]. The Gauss-Newton matrix for loss function $L(h(y))$ is derived in *Eq.* (5):

$$\mathbf{H}_L(h(y)) \approx \mathbf{G}_L(h(y)) = \mathbf{J}_h(y)^T \mathbf{J}_h(y). \tag{5}$$

The Hessian matrix $\mathbf{H}_L(h(y))$ and its approximation, the Gauss-Newton matrix $\mathbf{G}_L(h(y))$ are both square matrices of dimension $((|U| \times |I|) \times f) \times ((|U| \times |I|) \times f)$, while the Jacobian matrix $\mathbf{J}_h(y)$ corresponds to the model output $h(y)$ with the dimension of $(|R_K| \times (|U| \times |I|) \times f)$. These matrices are associated with the loss function $L(h(y))$ and its internal representation. The Jacobian is computed as follows:

$$\mathbf{J}_h(y) = \left( \frac{\partial}{\partial y} h_{(u,i)}(y) \Big|_{(u,i) \in K} \right). \tag{6}$$

*3.3 Hessian-Vector Product*

Due to limitations in computational and storage resources, most second-order LFA models employ the conjugate gradient method to solve *Eq.* (1) iteratively, which circumvents direct manipulation of the Hessian matrix's inverse. Furthermore, during each inner conjugate gradient iteration, these second-order-based models compute the Hessian-vector product through two Jacobian-vector products, thereby eliminating the need to store the curvature matrix. The Hessian-vector calculation mechanism is presented as follows: The Hessian-vector product computation procedure is defined by:

$$\mathbf{G}_L(h(y))v = R\{g_L(h(y))\} = \mathbf{J}_h(y)^T \mathbf{J}_h(y)v, \tag{7}$$

where $R\{\cdot\}$ stands for the $R$-operator, $v = vec(v_U, v_I) \in \mathbb{R}^{(|U| \times |I|) \times f}$ is the conjugate direction vector used in each conjugate gradient iteration, and $\mathbf{G}_L(h(y))v$ represents the Hessian-vector product. Here, $v_U \in \mathbb{R}^{|U| \times f}$ and $v_I \in \mathbb{R}^{|I| \times f}$ are the CG vectors corresponding to $Y_U$ and $Y_I$, respectively.

Actually, we do not explicitly manipulate the Gauss-Newton matrix $\mathbf{G}_L(h(y))$. Instead, we keep the Jacobian matrix $\mathbf{J}_h(y)$ and obtain the product with the Gauss-Newton matrix by performing two Jacobian-vector multiplications during each conjugate gradient iteration. Formally, the element-wise Jacobian-vector $\mathbf{J}_h(y)v$ can be expressed as:

$$\mathbf{J}_h(x)v = R\{h(x)\} = \left( \left( \sum_{d=1}^{f} (v_{u,d} x_{i,d} + x_{u,d} v_{i,d}) \right) \Big|_{(u,i) \in R_K} \right). \tag{8}$$

where $v_{u,d}$ denotes the $d$-th entry of the user-specific sub-vector $v_u$, and $v_{i,d}$ refers to the corresponding element in the item-specific sub-vector $v_i$. Here, $v_u$ and $v_i$ denote the $u$-th and $i$-th row vector of $v$, respectively.

*3.3 Integrating $L_2$ Regularization*

According to [8, 17], the objective function $E(y)$ includes loss function $L(h(y))$ and an $L_2$ regularization term $R(y)$, i.e., $E(y) = L(h(y)) + R(y)$. Consequently, the Hessian matrix of $E(y)$ decomposes as:

$$\mathbf{H}_E(y) \approx \mathbf{H}_L(h(y)) + \mathbf{H}_R(y), \tag{9}$$

where $\mathbf{H}_R(y)$ denotes the Hessian matrix of $R(y)$.

*3.4 Adaptive Cubic Regularization Newton Method*

Cubic regularization Newton method aims to minimize $T_M(y)$ as mentioned in *Eq.* (3). In the CG solver framework, *Eq.* (3) can be reformulated as follows:

$$\arg\min_{\Delta y} g_E(y) + (\mathbf{H}_E(y) + M \|\Delta y\| \mathbf{I}) \Delta x \leq \tau \mathbf{I}, \tag{10}$$

However, directly incorporating $M\|\Delta y\|$ as a damping term poses significant practical implementation challenges [15, 16]. Hence, a reasonable approach is to replace it with an approximate form, i.e., $M\|\Delta y\| \approx M\|g_E(y)\|$.

Building upon the preceding analysis, we can derive the element-wise Hessian-vector product associated with $E(y)$, incorporating the adaptive cubic regularization term $M\|g_E(y)\|$ as follows

$$c_E(y) \approx \left(\mathbf{G}_L(h(y)) + \mathbf{H}_R(y) + M\|g_E(y)\|\right)v$$

$$= \begin{cases} \text{For each } u \in U, \ d=1 \sim f: \\ c_E(y_u)_d = \left(\sum_{i \in R_{Ku}}\left(y_{u,d}\sum_{d=1}^{f}(v_{u,d}y_{i,d} + y_{u,d}v_{i,d})\right)\right) \\ \qquad + \lambda v_{u,d}|R_{Ku}| + M\|g_E(y)\| \\ \text{For each } i \in I, \ d=1 \sim f: \\ c_E(y_i)_d = \left(\sum_{u \in R_{Ki}}\left(y_{i,d}\sum_{d=1}^{f}(y_{u,d}v_{i,d} + v_{u,d}y_{i,d})\right)\right) \\ \qquad + \lambda v_{i,d}|R_{Ki}| + M\|g_E(y)\| \end{cases} \quad (11)$$

Let $c_E(y) \in \mathbb{R}^{(|U|\times|I|)\times f}$ denote the Hessian-vector product at each CG iteration. The element $c_E(y_u)_d$ refers to the $d$-th element of sub-vector $y_u$, and similarly, $c_E(y_i)_d$ corresponds to the $d$-th component of $i$-th sub-vector $y_i$. Here, $R_{Ku}$ and $R_{Ki}$ indicate the observed data associated with user $u$ and item $i$, respectively. The initial conjugate vector $v^0$ is set to the negative gradient of $E(y)$, i.e., $v^0 = -g_E(y)$.

*3.3 Update Rule*

After multiple CG iterations, the CG solver is going to output the increment $\Delta y \in \mathbb{R}^{(|U|\times|I|)\times f}$. The update rule at $t$-th training epoch is as given as follows:

$$y^{t+1} = y^t + \Delta y^t. \qquad (12)$$

*3.4 Algorithm Analysis*

Based on the above analysis, the ACRSLF model requires $O((|U|+|I|)\times f)$ storage. The computational bottleneck lies in the CG iterations per training epoch, not in the gradient computation or parameter updates. Hence, the time complexity for ACRSLF is $\Theta(T \times T_{CG} \times |R_K| \times f)$, where $T$ and $T_{CG}$ denote the maximum training iterations and CG iterations, respectively.

# 4 Experiments

*4.1 General Settings*

*Datasets.* There are two open-source HDI matrix datasets adopted in this section, including Yelp [105] and MovieLens 1M [99].

Table 1 The details of the datasets

| No. | Name | $|U|$ | $|I|$ | $|K|$ | Density |
|---|---|---|---|---|---|
| D1 | Yelp | 15,400 | 1,000 | 365,804 | 2.37% |
| D2 | ML-1M | 6,040 | 3,952 | 1,000,209 | 4.19% |

*Evaluation Metric.* To assess how effectively the LFA model captures low-rank structures, we adopt <u>r</u>oot <u>m</u>ean <u>s</u>quare <u>e</u>rror (RMSE) [2, 4, 7, 10, 17] as the primary evaluation

$$RMSE = \sqrt{\sum_{r_{u,i} \in \Lambda}\left(r_{u,i} - \sum_{d=1}^{f}y_{u,d}y_{i,d}\right)^2 \Big/ |\Lambda|} \qquad (13)$$

where $\Lambda$ denotes the evaluation set.

*Competitors.* The proposed ACRSLF model is compared to the state-of-the-art optimizer-based LFA models as follows:
- M1: SGD-M-based LFA model (M1) [1];
- M2: Adam-based LFA model (M2) [29];
- M3: SLF model [38];
- M4: ACRSLF model (M4), which is proposed in this paper.

*Experimental Setup.* The implementation details of benchmark datasets and comparative methods are configured as follows:

- *Environment Configuration:* Executed on an Intel Core i9-13905H platform (5.4 GHz, 32GB RAM) with Windows 11 Pro, utilizing OpenJDK 11 LTS runtime environment.
- *Hyperparameter Optimization:* Latent space dimension *f*=20, with matrix elements initialized by uniform sampling *U*(0, 0.004). M1, M2, and M3 hyperparameters follow original specifications [1, 29, 37], while M4 inherits M3's configuration with modified damping strength.
- *Terminate Condition:* Maximum 500 training epochs with early stopping and the patience=10 epochs).

*4.2 Comparison Results*

The table 2 presents the comparison results, revealing the following key observations:

Table 2 The details of the datasets

| Datasets | Model | RMSE | Time (Sec) | Epoch |
|---|---|---|---|---|
| D1 | M1 | 0.98877 | 27.332 | 393 |
|  | M2 | 0.98864 | 41.009 | 160 |
|  | M3 | 0.98835 | **9.318** | 74 |
|  | M4 | **0.98834** | 15.521 | **17** |
| D2 | M1 | 0.85484 | 21.657 | 457 |
|  | M2 | 0.85540 | 36.083 | 184 |
|  | M3 | 0.85427 | 23.270 | 105 |
|  | M4 | **0.85423** | **18.180** | **56** |

(1) **The proposed ACRSLF can converge efficiently while demonstrating a superior convergence rate and maintaining comparable prediction accuracy for handling missing data in HDI matrices.** For instance, benchmarking on the D1 dataset, M4 achieves a RMSE of 0.98834, virtually indistinguishable from SLF's 0.98835 measurement. In terms of convergence speed as measured by training epochs, M4 requires only 17 epochs compared to M3's 74 epochs.

(2) **Incorporating curvature information enhances the LFA model's ability to capture low-rank representations than its counterparts.** For instance, on the D2 dataset, M1–M3 yield RMSE values of 0.85484, 0.85540, and 0.85427, while the proposed M4 achieves 0.85423, reflecting marginal improvements of 0.07%, 0.14%, and 0.005%, respectively.

# 4 Conclusion

This paper introduces a novel latent factor analysis approach named ACRSLF, which incorporates cubic regularization and dynamically adjusts its damping term during training. This adaptive mechanism notably improves convergence efficiency. Empirical results show that ACRSLF consistently outperforms three leading baselines in both representation accuracy and training speed. In future work, we plan to explore theoretical underpinnings of the proposed framework.

# 6 References

[1] X. Luo, Y. Zhou, Z. Liu, and M. Zhou, "Fast and Accurate Non-negative Latent Factor Analysis on High-dimensional and Sparse Matrices in Recommender Systems," IEEE Trans. Knowl. Data Eng., vol. 35, no. 4 pp. 3897-3911, Apr. 2023.
[2] M. Shang, Y. Yuan, X. Luo, and M. Zhou, "An α-β-divergence-generalized Recommender for Highly-accurate Predictions of Missing User Preferences," IEEE Trans. Cybern., vol. 52, no. 8, pp. 8006-8018, Aug. 2022.
[3] X. Luo, Y. Yuan, S. Chen, N. Zeng, and Z. Wang, "Position-Transitional Particle Swarm Optimization-Incorporated Latent Factor Analysis," IEEE Trans. Knowl. Data Eng., vol. 34, no. 8, pp. 3958–3970, Aug. 2022.
[4] D. Wu, Z. Li, Z. Yu, Y. He, and X. Luo, "Robust Low-rank Latent Feature Analysis for Spatio-Temporal Signal Recovery," IEEE Tran. Neural Netw. Learn. Comput., early access, DOI: 10.1109/TNNLS.2023.3339786.
[5] W. Yang, S. Li, and X. Luo, "Data Driven Vibration Control: A Review," IEEE/CAA J. Autom. Sinica, vol. 11, no. 9, pp. 1898-1917, Sep. 2024.
[6] D. Wu, M. Shang, X. Luo, and Z. Wang, "An L1-and-L2-norm-oriented Latent Factor Model for Recommender Systems," IEEE Trans. Neural Netw. Learn. Syst., vol. 33, no. 10, pp. 5775-5788, Oct. 2022.
[7] Y. Yuan, Q. He, X. Luo, and M. Shang, "A Multilayered-and-Randomized Latent Factor Model for High-Dimensional and Sparse Matrices," IEEE Trans. Big Data, vol. 8, no. 3, pp. 784-794, Jun. 2022.
[8] D. Wu, P. Zhang, Y. He, and X. Luo, "MMLF: Multi-Metric Latent Feature Analysis for High-Dimensional and Incomplete Data," IEEE Trans. Serv. Comput., vol. 17, no. 2, Mar./Apr. 2024.

[9] M. Chen, Y. Qiao, R. Wang, and X. Luo, "A Generalized Nesterov's Accelerated Gradient-Incorporated Non-negative Latent-factorization-of-tensors Model for Efficient Representation to Dynamic QoS Data," IEEE Trans. Emerg. Top. Comput. Intell., vol. 8, no. 3, pp. 2386-2400, Jun. 2024.
[10] X. Xu, M. Lin, X. Luo, and Z. Xu, "HRST-LR: A Hessian Regularization Spatio-Temporal Low Rank Algorithm for Traffic Data Imputation," IEEE Trans. Intell. Transp. Syst., vol. 24, no. 10, pp. 11001-11017, Oct. 2023.
[11] F. Bi, X. Luo, B. Shen, H. Dong, and Z. Wang, "Proximal Alternating-Direction-Method-of-Multipliers-Incorporated Nonnegative Latent Factor Analysis," IEEE/CAA J. Autom. Sinica, vol. 10, no. 6, pp. 1388-1406, Jun. 2023.
[12] M. Chen, C. He, and X. Luo, "MNL: A Highly-Efficient model for Large-scale Dynamic Weighted Directed Network Representation," IEEE Trans. Big Data, vol. 9, no. 3, pp. 889-903, Jun. 2023.
[13] X. Luo, M. Chen, H. Wu, Z. Liu, H. Yuan, and M. Zhou, "Adjusting Learning Depth in Non-negative Latent Factorization of Tensors for Accurately Modeling Temporal Patterns in Dynamic QoS Data," IEEE Trans. Autom. Sci. Eng., vol. 18, no. 4, pp. 2142-2155, Oct. 2021.
[14] F. Bi, T. He, and X. Luo, "Two-Stream Graph Convolutional Network-Incorporated Latent Feature Analysis," IEEE Trans. Serv. Comput., vol. 16, no. 4, pp. 3027-3042, Jul./Aug. 2023.
[15] Y. Yuan, X. Luo, and M. Zhou, "Adaptive Divergence-based Non-negative Latent Factor Analysis of High-Dimensional and Incomplete Matrices from Industrial Applications," IEEE Trans. Emerg. Top. Comput. Intell., vol. 8, no. 2, Apr. 2024.
[16] Y. Zhong, L. Jin, M. Shang, and X. Luo, "Momentum-incorporated Symmetric Non-negative Latent Factor Models," IEEE Trans. Big Data, vol. 8, no. 4, pp. 1096-1106, Aug. 2022.
[17] X. Luo, H. Wu, and Z. Li, "NeuLFT: A Novel Approach to Nonlinear Canonical Polyadic Decomposition on High-Dimensional Incomplete Tensors," IEEE Trans. Knowl. Data Eng., vol. 35, no. 6, pp. 6148-6166, Jun. 2023.
[18] P. Tang, and X. Luo, "Neural Tucker Factorization," IEEE/CAA J. Autom. Sinica, vol. 12, no.2, pp. 457-477, Feb. 2024.
[19] F. Bi, T. He, and X. Luo, "A Fast Nonnegative Autoencoder-based Approach to Latent Feature Analysis on High-Dimensional and Incomplete Data," IEEE Trans. Serv. Comput., vol. 17, no. 3, pp. 733-746, May/Jun. 2024.
[20] Z. Liu, G. Yuan, and X. Luo, "Symmetry and Nonnegativity-Constrained Matrix Factorization for Community Detection," IEEE/CAA J. Autom. Sinica, vol. 9, no. 9, pp.1691-1693, Sep. 2022.
[21] D. Wu, Y. He, and X. Luo, "A Graph-incorporated Latent Factor Analysis Model for High-dimensional and Sparse Data," IEEE Trans. Emerg. Top. Comput. Intell., vol. 11, no. 4, pp. 907-917, Oct./Dec. 2023.
[22] X. Luo, H. Wu, Z. Wang, J. Wang, and D. Meng, "A Novel Approach to Large-Scale Dynamically Weighted Directed Network Representation," IEEE Trans. Pattern Anal. Mach. Intell., vol. 44, no. 12, pp. 9756-9773, Dec. 2022.
[23] Hao Wu, X. Luo, and M. Zhou, "Advancing Non-negative Latent Factorization of Tensors with Diversified Regularizations," IEEE Trans. Serv. Comput., vol. 15, no. 3, pp. 1334-1344, May/Jun. 2022.
[24] A. Bahamou, D. Goldfarb, and Y. Ren, "A Mini-Block Fisher Method for Deep Neural Networks," in Int. Conf. Artif. Intell. Statist., Apr. 2023, pp. 9191-9220.
[25] X. Luo, Y. Zhou, Z. Liu, L. Hu, and M. Zhou, "Generalized Nesterov's Acceleration-Incorporated, Non-Negative and Adaptive Latent Factor Analysis," IEEE Trans. Serv. Comput., vol. 15, no. 5, pp. 2809–2823, Sep./Oct. 2022.
[26] W. Qin and X. Luo, "Asynchronous Parallel Fuzzy Stochastic Gradient Descent for High-Dimensional Incomplete Data," IEEE Trans Fuzzy Syst., vol. 32, no. 2, pp. 445-459, Feb. 2024.
[27] X. Luo, W. Qin, A. Dong, K. Sedraoui, and M. Zhou, "Efficient and High-quality Recommendations via Momentum-incorporated Parallel Stochastic Gradient Descent-Based Learning," IEEE/CAA J. Autom. Sinica, vol. 8, no. 2, pp. 402–411, Feb. 2021.
[28] W. Qin, X. Luo, and M. Zhou, "Adaptively-accelerated Parallel Stochastic Gradient Descent for High-Dimensional and Incomplete Data Representation Learning," IEEE Trans. Big Data, vol. 10, no. 1, pp. 92-107, Feb. 2024.
[29] X. Luo, M. Zhou, Z. Wang, Y. Xia, and Q. Zhu, "An Effective Scheme for QoS Estimation via Alternating Direction Method-Based Matrix Factorization," IEEE Trans. Serv. Comput., vol. 12, no. 4, pp. 503–518, Jul./Aug. 2019.
[30] D. P. Kingma and J. Ba, "Adam: A method for stochastic optimization," in Proc. Int. Conf. Learn. Representations, May 2015.
[31] X. Luo, J. Chen, Y. Yuan, and Z. Wang, "Pseudo Gradient-Adjusted Particle Swarm Optimization for Accurate Adaptive Latent Factor Analysis," IEEE Trans. Syst., Man, Cybern. Syst., vol. 54, no. 4, pp. 2213-2226, Apr. 2024.
[32] X. Liao, K. Hoang, and X. Luo, "Local Search-based Anytime Algorithms for Continuous Distributed Constraint Optimization Problems," IEEE/CAA J. Autom. Sinica, DOI: 10.1109/JAS.2024.124413.
[33] J. Chen, R. Wang, G. Yuan, and X. Luo, "A Differential Evolution-Enhanced Position-Transitional Approach to Latent Factor Analysis," IEEE Trans. Emerg. Top. Comput. Intell., vol. 7, no. 2, pp. 389-401, Apr. 2023.

[34] X. Luo, M. Zhou, S. Li, Y. Xia, Z. You, Q. Zhu, and H. Leung, "An efficient second-order approach to factorize sparse matrices in recommender systems," IEEE Trans. Ind. Informat., vol. 11, no. 4, pp. 946–956, Aug. 2015.

[35] J. Martens, "Deep learning via Hessian-free optimization," in Proc. 27th Int. Conf. Mach. Learn., Jun. 2010, pp. 735–742.

[36] X. Luo, M. Zhou, S. Li, Y. Xia, Z. You, Q. Zhu, and H. Leung, "Incorporation of Efficient Second-Order Solvers Into Latent Factor Models for Accurate Prediction of Missing QoS Data," IEEE Trans. Cybern., vol. 48, no. 4, pp. 1216-1228, Apr. 2018.

[37] S. Li, Z. You, H. Guo, X. Luo, and Z. Zhao, "Inverse-Free Extreme Learning Machine With Optimal Information Updating," IEEE Trans. Cybern., vol. 46, no. 5, pp. 1229-1241, May 2016.

[38] W. Li, R. Wang, X. Luo, M. Zhou, "A Second-Order Symmetric Non-Negative Latent Factor Model for Undirected Weighted Network Representation," IEEE Trans. Netw. Sci. Eng., vol. 10, no. 2, vol. 606-618, 2023.

[39] J. Wang, W. Li, and X. Luo, "A Distributed Adaptive Second-order Latent Factor Analysis Model," IEEE/CAA J. Autom. Sinica, early access, DOI: 10.1109/JAS.2024.124371.

[40] W. Li, X. Luo, H. Yuan, M. Zhou, "A Momentum-Accelerated Hessian-Vector-Based Latent Factor Analysis Model," IEEE Trans. Serv. Comput., vol. 16, no. 2, pp. 830-844, Mar./Apr. 2023.

[41] S. Li, M. Zhou, and X. Luo, "Modified Primal-Dual Neural Networks for Motion Control of Redundant Manipulators With Dynamic Rejection of Harmonic Noises," IEEE Trans. Neural Netw. Learn. Syst., vol. 29, no. 10, pp. 4791-1801, Oct. 2018.

[42] H. Wu, X. Luo, M. Zhou, M. Rawa, K. Sedraoui, and A. Albeshri, "A PID-incorporated Latent Factorization of Tensors Approach to Dynamically Weighted Directed Network Analysis," IEEE/CAA J. Autom. Sinica, vol. 9, no. 3, pp. 533–546, Mar. 2021.

[43] X. Luo, Z. Li, W. Yue, and S. Li, "A Calibrator Fuzzy Ensemble for Highly-Accurate Robot Arm Calibration," IEEE Trans. Neural Netw. Learn. Syst., early access, DOI: 10.1109/TNNLS.2024.3354080.

[44] Z. Li, S. Li, O. Bamasag, A. Alhothali, and X. Luo, "Diversified Regularization Enhanced Training for Effective Manipulator Calibration," IEEE Trans. Neural Netw. Learn. Syst., vol. 34, no. 11, pp. 8788-8790, Nov. 2023.

[45] S. Li, M. Zhou, X. Luo, and Z. You, "Distributed Winner-Take-All in Dynamic Networks," IEEE Trans. Autom. Control., vol. 62, no. 2, pp. 577-589, Feb. 2017.

[46] Z. Li, X. Luo, and S. Li, "Efficient Industrial Robot Calibration via a Novel Unscented Kalman Filter-Incorporated Variable Step-Size Levenberg-Marquardt Algorithm," IEEE Trans. Instrum. Meas., vol. 72, pp. 1-12, Apr. 2023.

[47] Z. Li, S. Li, and X. Luo, "Using Quadratic Interpolated Beetle Antennae Search to Enhance Robot Arm Calibration Accuracy," IEEE Robotics Autom. Lett., vol. 7, no. 4, pp. 12046-12053, Oct. 2022.

[48] L. Jin, S. Liang, X. Luo, and M. Zhou, "Distributed and Time-Delayed k-Winner-Take-All Network for Competitive Coordination of Multiple Robots," IEEE Trans. Cybern., vol. 53, no. 1, pp. 641-652, Jan. 2023.

[49] Z. Li, S. Li, and X. Luo, "A Novel Machine Learning System for Industrial Robot Arm Calibration," IEEE Trans. Circuits Syst. II Express Briefs, vol. 71, no. 4, pp. 2364-2368, Apr. 2024.

[50] L. Jin, Y. Li, X. Zhang, and X. Luo, "Fuzzy k-Winner-Take-All Network for Competitive Coordination in Multi-robot Systems," IEEE Trans. Fuzzy Syst., vol. 32, no. 4, pp. 2005-2016, Apr. 2024.

[51] Z. Li, S. Li, A. Francis, and X. Luo, "A Novel Calibration System for Robot Arm via An Open Dataset and A Learning Perspective," IEEE Trans. Circuits Syst. II Express Briefs, vol. 69, no. 12, pp. 5169-5173, Dec. 2022.

[52] Y. Yuan, X. Luo, M. Shang, and Z. Wang, "A Kalman-Filter-Incorporated Latent Factor Analysis Model for Temporally Dynamic Sparse Data," IEEE Trans. Cybern., vol. 53, no. 9, pp. 5788-5801, Sep. 2023.

[53] T. Chen, S. Li, Y. Qiao, and X. Luo, "A Robust and Efficient Ensemble of Diversified Evolutionary Computing Algorithms for Accurate Robot Calibration," IEEE Trans. Instrum. Meas., vol. 73, pp. 1-14, Feb. 2024.

[54] J. Li, X. Luo, Y. Yuan, and S. Gao, "A Nonlinear PID-Incorporated Adaptive Stochastic Gradient Descent Algorithm for Latent Factor Analysis," IEEE Trans. Netw. Sci. Eng., early access, Jun. 2023, DOI: 10.1109/TASE.2023.3284819.

[55] J. Chen, Y. Yuan, and X. Luo, "SDGNN: Symmetry-Preserving Dual-Stream Graph Neural Networks," IEEE/CAA J. Autom. Sinica, vol. 11, no. 7, pp. 1717-1719, Jul. 2024.

[56] Z. Liu, X. Luo, and M. Zhou, "Symmetry and Graph Bi-regularized Non-Negative Matrix Factorization for Precise Community Detection," IEEE Trans. Autom. Sci. Eng., vol. 21, no. 2, Apr. 2024.

[57] D. Wu, X. Luo, Y. He, and M. Zhou, "A Prediction-sampling-based Multilayer-structured Latent Factor Model for Accurate Representation to High-dimensional and Sparse Data," IEEE Trans. Neural Netw. Learn. Syst., vol. 35, no. 3, pp. 3845-3858, Mar. 2024.

[58] Z. Liu, Y. Yi, and X. Luo, "A High-Order Proximity-Incorporated Nonnegative Matrix Factorization-based Community Detector," IEEE Trans. Emerg. Top. Comput. Intell., vol. 7, no. 3, pp. 700-714, Jun. 2023.
[59] D. Wu, P. Zhang, Y. He, and X. Luo, "A Double-Space and Double-Norm Ensembled Latent Factor Model for Highly Accurate Web Service QoS Prediction," IEEE Tran. Serv. Comput., vol. 16, no. 2, pp. 802-814, Mar./Apr. 2023.
[60] X. Luo, Z. Liu, L. Jin, Y. Zhou, and M. Zhou, "Symmetric Non-negative Matrix Factorization-based Community Detection Models and Their Convergence Analysis," IEEE Trans. Neural Netw. Learn. Syst., vol. 33, no. 3, pp. 1203-1215, Mar. 2022.
[61] .D. Wu, X. Luo, M. Shang, Y. He, G. Wang, and X. Wu, "Data-Characteristic-Aware Latent Factor Model for Web Services QoS Prediction," IEEE Trans. Knowl. Data Eng., vol. 34, no. 6, pp. 2525-2538, Jun. 2022.
[62] X. Luo, L. Wang, P. Hu, and L. Hu, "Predicting Protein-Protein Interactions Using Sequence and Network Information via Variational Graph Autoencoder," IEEE/ACM Trans. Comput. Biol. Bioinform., vol. 20, no. 5, pp. 3182-3194, Sep./Oct. 2023.
[63] X. Shi, Q. He, X. Luo, Y. Bai, and M. Shang, "Large-scale and Scalable Latent Factor Analysis via Distributed Alternative Stochastic Gradient Descent for Recommender Systems," IEEE Trans. Big Data, vol. 8, no. 2, pp. 420-431, Apr. 2022.
[64] D. Wu, Q. He, X. Luo, and M. Shang, Y. He, and G. Wang, "A Posterior-neighborhood-regularized Latent Factor Model for Highly Accurate Web Service QoS Prediction," IEEE Trans. Serv. Comput., vol. 15, no. 2, pp. 793-805, Mar./Apr. 2022.
[65] Q. Jiang, D. Liu, H. Zhu, S. Wu, N. Wu, X. Luo, and Y. Qiao, "Iterative Role Negotiation via the Bi-level GRA++ with Decision Tolerance," IEEE Trans. Comput. Soc. Syst., early access, DOI: 10.1109/TCSS.2024.3409893.
[66] Y. Zhong, K. Liu, S. Gao, and X. Luo, "Alternating-Direction-Method of Multipliers-based Adaptive Nonnegative Latent Factor Analysis," IEEE Trans. Emerg. Top. Comput. Intell., early access, DOI: 10.1109/TETCI.2024.3420735.
[67] X. Luo, Y. Zhong, Z. Wang, and M. Li, "An Alternating-direction-method of Multipliers-Incorporated Approach to Symmetric Non-negative Latent Factor Analysis," IEEE Trans. Neural Netw. Learn. Syst., vol. 34, no. 8, pp. 4826-4840, Aug. 2023.
[68] D. Wu, Y.He, X. Luo, and M. Zhou, "A Latent Factor Analysis-based Approach to Online Sparse Streaming Feature Selection," IEEE Trans. Syst. Man Cybern. Syst., vol. 52, no. 11, pp. 6744-6758, Nov. 2022.
[69] W. Li, R. Wang, and X. Luo, "A Generalized Nesterov-Accelerated Second-Order Latent Factor Model for High-Dimensional and Incomplete Data," IEEE Trans. Neural Netw. Learn. Syst., early access, DOI: 10.1109/TNNLS.2023.3321915.
[70] Y. Yuan, J. Li, and X. Luo, "A Fuzzy PID-Incorporated Stochastic Gradient Descent Algorithm for Fast and Accurate Latent Factor Analysis," IEEE Trans. Fuzzy Syst., early access, Apr. 2024, DOI: 10.1109/TFUZZ.2024.3389733.
[71] L. Jin, L. Wei, and S. Li, "Gradient-Based Differential Neural-Solution to Time-Dependent Nonlinear Optimization," IEEE Trans. Autom. Control., vol. 68, no. 1, pp. 620-627, Jan. 2023.
[72] N. Zeng, X. Li, P. Wu, H. Li, and X. Luo, "A Novel Tensor Decomposition-based Efficient Detector for Low-altitude Aerial Objects with Knowledge Distillation Scheme," IEEE/CAA J. Autom. Sinica, vol. 11, no. 2, pp. 487-501, Feb. 2024.
[73] J. Li, F. Tan, C. He, Z. Wang, H. Song, P. Hu, and X. Luo, "Saliency-Aware Dual Embedded Attention Network for Multivariate Time-Series Forecasting in Information Technology Operations," IEEE Trans. Ind. Informatics, vol. 20, no. 3, pp. 4206-4217, Mar. 2024.
[74] L. Hu, Y. Yang, Z. Tang, Y. He, and X. Luo, "FCAN-MOPSO: An Improved Fuzzy-based Graph Clustering Algorithm for Complex Networks with Multi-objective Particle Swarm Optimization," IEEE Trans. Fuzz. Syst., vol. 31, no. 10, pp. 3470-3484, Oct. 2023.
[75] L. Chen and X. Luo, "Tensor Distribution Regression based on the 3D Conventional Neural Networks," IEEE/CAA J. Autom. Sinica, vol. 7, no. 3, pp. 700-714, Jul. 2023.
[76] W. Qin, H. Wang, F. Zhang, J. Wang, X. Luo, and T Huang, "Low-Rank High-Order Tensor Completion with Applications in Visual Data," IEEE Trans. Image Process, vol. 31, pp. 2433-2448, Mar. 2022.
[77] Q. Wang, X. Liu, T. Shang, Z. Liu, H. Yang, and X. Luo, "Multi-Constrained Embedding for Accurate Community Detection on Undirected Networks," IEEE Trans. Netw. Sci. Eng., vol. 9, no. 5, Sept./Oct. 2022.
[78] Y. Song, Z. Zhu, M. Li, G. Yang, and X. Luo, "Non-negative Latent Factor Analysis-Incorporated and Feature-Weighted Fuzzy Double c-Means Clustering for Incomplete Data," IEEE Trans. Fuzz. Syst., vol. 30, no. 10, Oct. 2022.
[79] J. Chen, X. Luo, and M. Zhou, "Hierarchical Particle Swarm Optimization-incorporated Latent Factor Analysis for Large-Scale Incomplete Matrices," IEEE Trans. Big Data, vol. 8, no. 6, Dec. 2022.

[80] L. Hu, X. Pan, Z. Tang, and X. Luo, "A Fast Fuzzy Clustering Algorithm for Complex Networks via a Generalized Momentum Method," IEEE Trans. Fuzz. Syst., vol. 30, no. 9, Sep. 2022.
[81] D. Cheng, J. Huang, S. Zhang, X. Zhang, and X. Luo, "A Novel Approximate Spectral Clustering Algorithm with Dense Cores and Density Peaks," IEEE Trans. Syst. Man Cybern. Syst., vol. 52, no. 4, pp. 2348-2360, Apr. 2022.
[82] F. Zhang, L. Jin, and X. Luo, "Error-Summation Enhanced Newton Algorithm for Model Predictive Control of Redundant Manipulators," IEEE Trans. Ind. Electron., vol. 70, no. 3, pp. 2800-2811, Mar. 2023.
[83] L. Wei, L. Jin, and X. Luo, "A Robust Coevolutionary Neural-Based Optimization Algorithm for Constrained Nonconvex Optimization," IEEE Trans. Neural Netw. Learn. Syst., vol. 35, no. 6, pp. 7778-7791, Jun. 2024.
[84] Z. Xie, L. Jin, X. Luo, B. Hu, and S. Li, "An Acceleration-Level Data-Driven Repetitive Motion Planning Scheme for Kinematic Control of Robots With Unknown Structure," IEEE Trans. Syst. Man Cybern. Syst., vol. 52, no. 9, pp. 5679-5691, Sep. 2022.
[85] X. Xiao, Y. Ma, Y. Xia, M. Zhou, X. Luo, X. Wang, X. Fu, W. Wei, and N. Jiang, "Novel Workload-aware Approach to Mobile User Reallocation in Crowded Mobile Edge Computing Environment," IEEE Trans. Intell. Transp. Syst., vol. 23, no. 7, pp. 8846-8856, Jul. 2022.
[86] Z. Xie, L. Jin, X. Luo, Z. Sun, and M. Liu, "RNN for Repetitive Motion Generation of Redundant Robot Manipulators: An Orthogonal Projection Based Scheme," IEEE Trans. Neural Learn. Syst., vol. 33, no. 2, pp. 615-628, Jul. 2022.
[87] L. Wei, L. Jin, and X. Luo, "Noise-Suppressing Neural Dynamics for Time-Dependent Constrained Nonlinear Optimization With Applications," IEEE Trans. Syst. Man Cybern. Syst., vol. 52, no. 10, pp. 6139-6150, Oct. 2022.
[88] J. Yan, L. Jin, X. Luo, and S. Li, "Modified RNN for Solving Comprehensive Sylvester Equation with TDOA Application," IEEE Trans. Neural Netw. Learn. Syst., early access, DOI: 10.1109/TNNLS.2023.3263565.
[89] Z. Wang, Y. Liu, X. Luo, J. Wang, C. Gao, D. Peng, and W. Chen, "Large-Scale Affine Matrix Rank Minimization with a Novel Nonconvex Regularizer," IEEE Trans. Neural Netw. Learn. Syst., vol. 33, no. 9, pp. 4661-4675, Sep. 2022.
[90] L. Jin, Y. Qi, X. Luo, S. Li, and M. Shang, "Distributed Competition of Multi-Robot Coordination under Variable and Switching Topologies," IEEE Trans. Autom. Sci. Eng., vol. 19, no. 4, pp. 3575-3586, Oct. 2022.
[91] L. Hu, S. Yang, X. Luo, H. Yuan, and M. Zhou, "A Distributed Framework for Large-scale Protein-protein Interaction Data Analysis and Prediction Using MapReduce," IEEE/CAA J. Autom. Sinica, vol. 9, no. 1, pp. 160-172, Jan. 2022.
[92] L. Jin, X. Zheng, and X. Luo, "Neural Dynamics for Distributed Collaborative Control of Manipulators with Time Delays," IEEE/CAA J. Autom. Sinica, vol. 9, no. 5, pp. 854-863, Jan. 2022.
[93] Y. Qi, L. Jin, X. Luo, and M. Zhou, "Recurrent Neural Dynamics Models for Perturbed Nonstationary Quadratic Programs: A Control-theoretical Perspective," IEEE Trans. Neural Netw. Learn. Syst., vol. 33, no. 3, pp. 1216-1227, Mar. 2022.
[94] Y. Qi, L. Jin, X. Luo, Y. Shi, and M. Liu, "Robust k-WTA Network Generation, Analysis, and Applications to Multi-Agent Coordination," IEEE Trans. Cybern., vol. 52, no. 8, pp. 8515-8527, Aug. 2022.
[99] F. M. Harper and J. A. Konstan, "The movielens datasets: History and context," ACM Trans. Interact. Intell. Syst., vol. 5, no. 4, pp. 1-19, Jan. 2016.
[95] H. Wu and X. Luo, "Discovering Hidden Pattern in Large-scale Dynamically Weighted Directed Network via Latent Factorization of Tensors," in Proc. 17th IEEE Int. Conf. Autom. Sci. Eng., 2021, pp. 1533-1538.
[96] I. Cantador, P. Brusilovsky, and T. Kuflik, "Second workshop on information heterogeneity and fusion in recommender systems," in Proc. 5th ACM Conf. Rec. Syst., Oct. 2011, pp. 387–388.
[97] H. Wu, X. Luo, and M. Zhou, "Neural Latent Factorization of Tensors for Dynamically Weighted Directed Networks Analysis," in Proc. 2021 IEEE Int. Conf. Syst. Man Cybern., 2021, pp. 3061-3066.
[98] T. T. Nguyen, F. M. Harper, L. Terveen, and J. A. Konstan, "User Personality and User Satisfaction with Recommender Systems," Inf. Syst. Front., vol. 20, pp. 1173–1189, 2018.
[99] H. Wu, X. Luo, and M. Zhou, "Instance-Frequency-Weighted Regularized, Nonnegative and Adaptive Latent Factorization of Tensors for Dynamic QoS Analysis," in Proc. 2021 IEEE Int. Conf. Web Serv., 2021, pp. 560-568.
[100] M. Sharma, F. M. Harper, and G. Karypis, "Learning from Sets of Items in Recommender Systems," ACM Trans. Interact. Intell. Syst., vol. 9, no. 4, pp. 1-16, Jul. 2019.
[101] H. Wu, Y. Xia, and X. Luo, "Proportional-Integral-Derivative-Incorporated Latent Factorization of Tensors for Large-Scale Dynamic Network Analysis," in Proc. 2021 China Autom. Congress, 2021, pp. 2980-2984.
[102] Q. Wang and H. Wu, "Dynamically Weighted Directed Network Link Prediction Using Tensor Ring Decomposition," in 2024 27th Int. Conf. Comput. Supported Cooperatvie Work Design, 2024, pp. 2864-2869

[103] H. Wu, X. Wu, and X. Luo, "Dynamic Network Representation Based on Latent Factorization of Tensors," Springer, 2023.

[104] X. Luo, H. Wu, H. Yuan, and M. Zhou, "Temporal Pattern-Aware QoS Prediction via Biased Non-Negative Latent Factorization of Tensors," IEEE Trans. Cybern., vol. 50, no. 5, pp. 1798-1809, May 2020.

[105] "Yelp Open Dataset", https://business.yelp.com/data/resources/open-dataset/.

[106] B. A. Pearlmutter, "Fast exact multiplication by the Hessian," Neural computation, vol. 6, no. 1, pp. 147-160, Jan. 1994.